\documentclass[10pt,twocolumn,letterpaper]{article}

\usepackage{cvpr}
\usepackage{times}
\usepackage{epsfig}
\usepackage{graphicx}
\usepackage{amsmath}
\usepackage{amssymb}
\usepackage{threeparttable}
\usepackage{multirow}
\usepackage{float}
\usepackage{multicol}
\usepackage{graphicx}
\usepackage{array,multirow}
\usepackage{setspace}
\usepackage{eucal}
\usepackage[bookmarks=false]{hyperref}
\usepackage{xcolor}
\hypersetup{
  colorlinks,
  linkcolor={red!70!black},
  citecolor={blue!50!black},
  urlcolor={blue!80!black}
}
 \cvprfinalcopy 

\def\cvprPaperID{474} 

\newcommand{\cutsectionup}{\vspace*{-0.12in}}
\newcommand{\cutsectiondown}{\vspace*{-0.05in}}

\ifcvprfinal\fi
\begin{document}

\begin{multicols}{2}
\title{Feedback Networks\vspace{-3pt}}

\author{Amir R. Zamir$^{1,3}$\thanks{Authors contributed equally.} \;\; Te-Lin Wu$^{1}$\footnotemark[1] \;\; Lin Sun$^{1,2}$ \;\;  William B. Shen$^{1}$ \;\; Bertram E. Shi\vspace{3pt}$^{2}$ \\ Jitendra Malik$^{3}$ \;\; Silvio Savarese$^{1}$\vspace{7pt}\\ 
$^1$ Stanford University \;\; $^2$ HKUST  \;\;  
$^3$ University of California, Berkeley\\ \vspace{4pt}
\textcolor{blue}{\url{http://feedbacknet.stanford.edu/}\vspace{-8pt}}
}



\maketitle
\end{multicols}

\begin{abstract}
Currently, the most successful learning models in computer vision are based on learning successive representations followed by a decision layer. This is usually actualized through feedforward multilayer neural networks, e.g. ConvNets, where each layer forms one of such successive representations. However, an alternative that can achieve the same goal is a feedback based approach in which the representation is formed in an iterative manner based on a feedback received from previous iteration's output.  

We establish that a feedback based approach has several core advantages over feedforward: it enables making early predictions at the query time, its output naturally conforms to a hierarchical structure in the label space (e.g. a taxonomy), and it provides a new basis for Curriculum Learning. We observe that feedback develops a considerably different representation compared to feedforward counterparts, in line with the aforementioned advantages. We present a general feedback based learning architecture, instantiated using existing RNNs, with the endpoint results on par or better than current feedforward networks and the addition of the above advantages. 
\vspace{-19pt}
\end{abstract}
\section{Introduction} \label{sec:intro}
Feedback is defined to occur when the (full or partial) output of a system is routed back into the input as part of an iterative cause-and-effect process~\cite{ford1999modeling}. 
Utilizing feedback is a strong way of making predictions in various fields, ranging from control theory to psychology~\cite{lee1967foundations,parlos1994application,ashford1983feedback}. Employing feedback connections is also heavily exercised by the brain suggesting a core role for it in complex cognition~\cite{holling1973resilience,rust2010selectivity,rust2010selectivity,cichy2014resolving,lee2003hierarchical}. In this paper, we show that a feedback based learning approach has several advantages over the commonly employed feedforward paradigm making it a worthwhile alternative. These advantages (elaborated below) are mainly attributed to the fact that the final prediction is made in an iterative, rather than one-time, manner along with an explicit notion of the thus-far output per iteration.  

\begin{figure}
  \centering
  \centerline{\includegraphics[width=1\columnwidth]{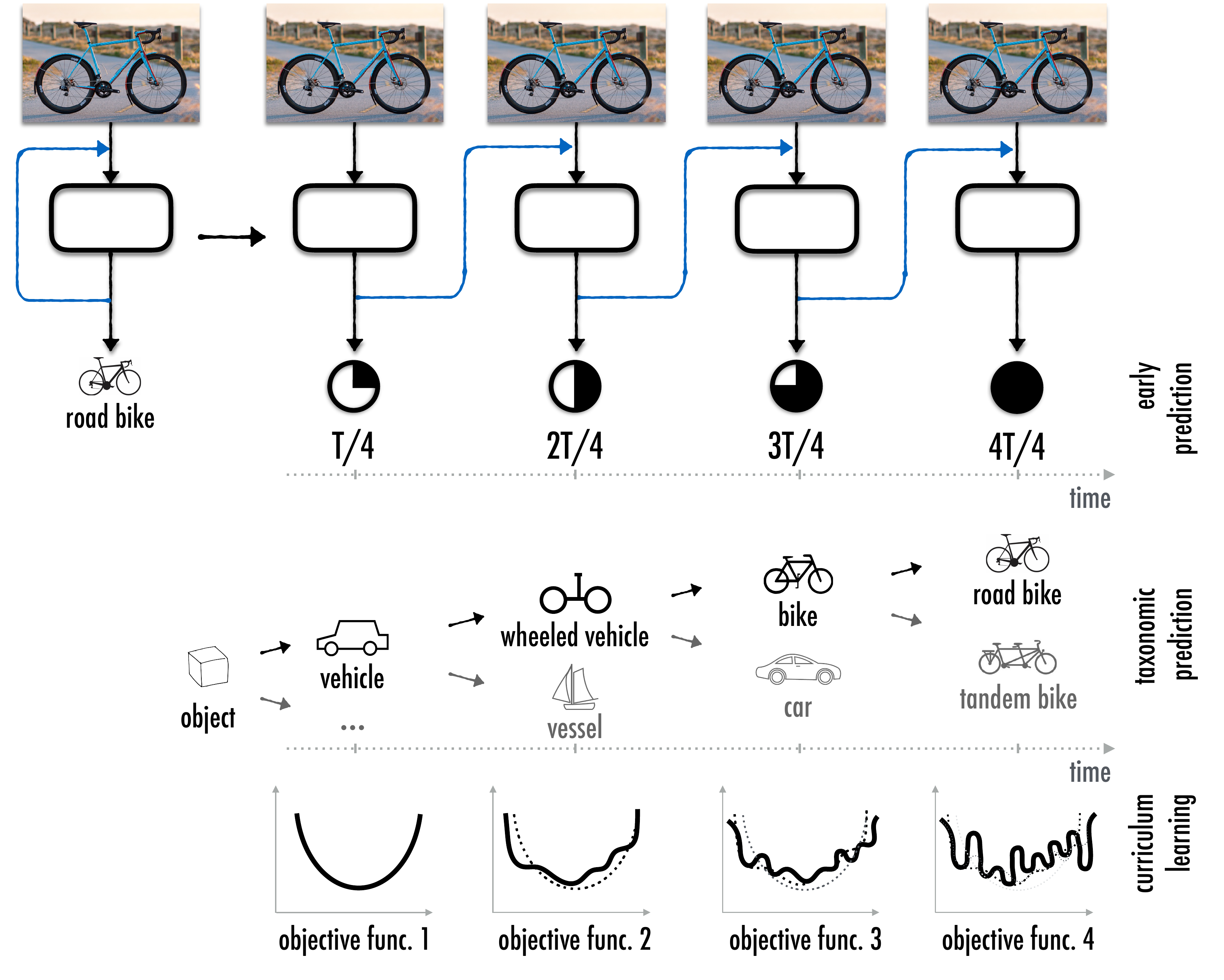}}
  \caption{\footnotesize \textbf{A feedback based learning model.} The basic idea is to make predictions in an iterative manner based on a notion of the thus-far outcome. This provides several core advantages: I. enabling early predictions (given total inference time $T$, early predictions are made in fractions of $T$); II. naturally conforming to a taxonomy in the output space; and III. better grounds for curriculum learning.} \label{fig:pull}
\end{figure}

\textbf{Early Predictions:} One advantage is providing estimations of the output in a fraction of the total inference time. This is schematically illustrated in Fig.~\ref{fig:pull}. This property is a result of iterative inference and is in contrast to feedforward where a one-time output is provided only when the signal reaches the end of the network. This is of particular importance in practical scenarios, such as robotics or autonomous driving; e.g. imagine a self driving car that receives a cautionary heads up about possibly approaching a pedestrian on a highway, without needing to wait for the final definite output. Such scenarios are abundant in practice as usually time is crucial and limited computation resources can be reallocated based on early predictions on-the-fly, given a proper uncertainty measure, such as Minimum Bayes Risk~\cite{kumar2004minimum}. 

\textbf{Taxonomy Compliance:} Another advantage is making predictions that naturally conform to a hierarchical structure in the output space, e.g. a taxonomy, even when not trained using the taxonomy. The early predictions of the feedback model conform to a coarse classification, while the later iterations further decompose the coarse class into finer classes. This is illustrated in Fig.\ref{fig:pull}. This is again due to the fact that the predictions happen in an iterative manner coupled with a \emph{coarse-to-fine representation}. The coarse-to-fine representation is naturally developed as the network is forced to make a prediction as early as the first iteration and iteratively improve it in all following iterations. 

\textbf{Episodic Curriculum Learning:} The previous advantage is closely related to the concept of Curriculum Learning~\cite{bengio2009curriculum}, where gradually increasing the complexity of the task leads to a better training~\cite{elman1993learning,bengio2009curriculum,krueger2009flexible}. For non-convex training criteria (such as in ConvNets), a curriculum is known to assist with finding better minima; in convex cases, it improves the convergence speed~\cite{bengio2009curriculum}. 

As prediction in a feedforward network happens in a one-time manner, a curriculum has to be enforced through feeding the training data in an order based on complexity (i.e. first epochs formed of easy examples and later the hard ones). In contrast, the predictions in a feedback model are made in an iterative form, and this enables enforcing a curriculum \emph{through the episodes of prediction for one query}. We call this \emph{Episodic Curriculum Learning}. In other words, sequential easy-to-hard decisions can be enforced for one datapoint (e.g. training the early episodes to predict the species and the later episodes the particular breed). Hence, any taxonomy can be used as a curriculum strategy. 




In our model, we define feedback based prediction as a recurrent (weight) shared operation, where at each iteration the output is estimated and passed onto the next iteration through a hidden state. The next iteration then makes an updated prediction using the shared operation and received hidden state. It is crucial for the hidden state to carry a direct notion of output, otherwise the entire system would be a feedforward pass realized through a recurrent operation~\cite{liang2015recurrent}. Therefore, we train the network to make a prediction at each iteration by backpropagating the loss in all iterations. We present a generic architecture for such networks, instantiated simply using existing RNNs, and empirically prove the aforementioned advantages on various datasets. 
Though we show that the feedback approach achieves competent final results, the primary goal of this paper is to establish the aforementioned conceptual properties, rather than optimizing for endpoint performance on any benchmark. The developed architectures and pre-trained models are available at \textcolor{blue}{\url{http://feedbacknet.stanford.edu/}}. 

\section{Related Work}

There is a notable amount of prior research in machine learning~\cite{tu2008auto,pinheiro2014recurrent,tompson2015efficient,oberweger2015training,veit2016residual,greff2016highway,gregor2010learning,wang2016learning,socher2011parsing,gkioxari2016chained,byeon2015scene,socher2012convolutional} and neuroscience~\cite{gilbert2007brain,hupe1998cortical,wyatte2012limits} that have commonalities with feedback based learning. We provide a categorized overview of some of the most related works. 

Conventional feedforward networks, e.g. AlexNet~\cite{AlexNet}, do not employ recurrence or feedback mechanisms. A number of recent successful methods used recurrence-inspired mechanisms in feedforward models. An example is ResNet~\cite{he2015deep}, introducing parallel residual connections, as well as hypernetworks~\cite{ha2016hypernetworks}, highway networks~\cite{srivastava2015highway}, stochastic depth~\cite{huang2016deep}, RCNN~\cite{liang2015recurrent}, GoogLeNet~\cite{szegedy2015going}. These methods are still feedforward as iterative injection of the thus-far output into the system is essential for forming a proper feedback. We empirically show that this requirement, besides recurrence, is indeed critical (Table~\ref{table:sequencer_effect}).

Several recent methods explicitly employed feedback connections~\cite{carreira2015human,belagiannis2016recurrent,xingjian2015convolutional,li2015iterative,liao2016bridging,karpathy2015deep} with promising results for their task of interest. The majority of these methods are either task specific and/or model temporal problems. Here we put forth and investigate the core advantages of a general feedback based inference. We should also emphasize that feedback in our model is always in the hidden space. This allows us to develop generic feedback based architectures without the requirement of task-specific error-to-input functions~\cite{carreira2015human} (See  {\href{http://feedbacknet.stanford.edu/supplementary_material}{supplementary material}} (Sec.~2) for more discussions). Stacked inference methods are also another group of related works~\cite{wolpert1992stacked,weiss2010structured,tu2008auto,toshev2014deeppose,ramakrishna2014pose}. Unlike the method studied here, many of them treat their outputs in isolation and/or do no employ weight sharing.

Another family of methods use feedback like mechanisms for spatial attention~\cite{xu2015show,cao2015look,mnih2014recurrent,mnih2014recurrent,wang2014attentional,stollenga2014deep}. This is usually used for better modeling of long term dependencies, computational efficiency, and spatial localization. 
Lastly, it is worth noting that Curriculum Learning~\cite{elman1993learning,krueger2009flexible,bengio2009curriculum} and making predictions on a taxonomy~\cite{hu2015learning,song2010taxonomic,deng2014large,ding2015probabilistic,koller2009probabilistic} are well investigated in the literature, though none provided a feedback based approach which is our focus.


\section{Feedback Networks}
\begin{figure}
  \centering
  \centerline{\includegraphics[width=1\columnwidth]{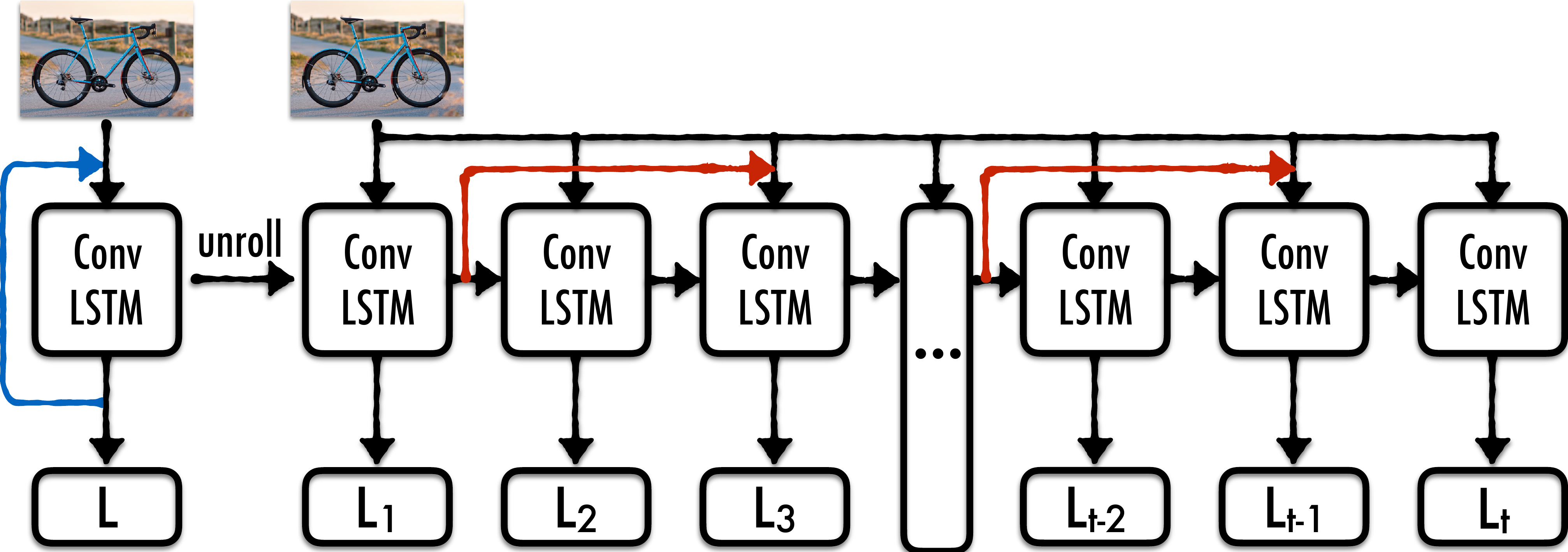}}
  \caption{\footnotesize \textbf{Illustration of our core feedback model and skip connections (shown in red) when unrolled in time.} `ConvLSTM' and `L' boxes represent convolutional operations and iteration losses, respectively.}
  \label{fig:skipConnect}
\end{figure}

Feedback based prediction has two requirements: (1) iterativeness and (2) rerouting a notion of posterior (output) back into the system in each iteration. We instantiate this by adopting a convolutional recurrent neural network model and connecting the loss to each iteration. The overall process can be summarized as: the image undergoes a shared convolutional operation repeatedly and a prediction is made at each time; the recurrent convolutional operations are trained to produce the best output at each iteration given a hidden state that carries a direct notation of thus-far output. This is depicted in Fig.~\ref{fig:skipConnect}.

\subsection{Convolutional LSTM Formulation} 
\label{functions_section}
In this section, we share the details of our feedback model which is based on stacking a flexible variant of ConvLSTM~\cite{xingjian2015convolutional} modules that essentially replace the operations in an LSTM~\cite{lstm} cell with convolutional structures\footnote{See  {\href{http://feedbacknet.stanford.edu/supplementary_material}{supplementary material}} (Sec.~7) for a discussion on alternatives to LSTM for this purpose, including GRU, vanilla RNN, and ablated LSTM.}. An LSTM cell uses hidden states to pass information through iterations. We briefly describe the connections between stacked ConvLSTMs and the gates in them:

We parametrize the temporal order (i.e. iterations) with time $t = 0,1,...,T$ and spatial order of a ConvLSTM module in the stack with depth $d = 0,1,...,D$. At depth $d$ and time $t$, the output of a ConvLSTM module is based on spatial input ($\mathbf{X}^{d-1}_t$), temporal hidden state input ($\mathbf{H}^d_{t-1}$), and temporal cell gate input ($\mathbf{C}^d_{t-1}$). 

To compute the output of a ConvLSTM module, the input gate $i^d_t$ and forget gate $f^d_t$ are used to control the information passing between hidden states:
\begin{eqnarray}\label{eqn:convLSTM}
\begin{aligned}
i^d_t &= \sigma(W_{d,xi}(\mathbf{X}^{d-1}_t) + W_{d,hi}(\mathbf{H}^d_{t-1})),\\
f^d_t &= \sigma(W_{d,xf}(\mathbf{X}^{d-1}_t) + W_{d,hf}(\mathbf{H}^d_{t-1})),\\
\end{aligned}
\end{eqnarray}
\newline
where $\sigma$ is sigmoid function. $W$ is a set of feedforward convolutional operations applied to $\mathbf{X}$ and $\mathbf{H}$. Here $W$ is parametrized by $d$ but not $t$ since the weights of convolutional filters are shared in the temporal dimension. The architecture of $W$ is a design choice and is the primary difference between our ConvLSTM module and Xingjian et al.~\cite{xingjian2015convolutional} as we use multilayer convolutional operations for $W$ with flexibility of including residual connections. The depth of $W$ (i.e. the physical depth of a ConvLSTM module) is discussed in Sec. \ref{feedback_type_section}.
\newline
\newline
The cell gate $\mathbf{C}^d_t$ is computed as follows:
\begin{eqnarray}\label{eqn:convLSTM}
\begin{aligned}
\tilde{C^d_t} &= tanh(W_{d,xc} (\mathbf{X}^{d-1}_t) + W_{d,hc}( \mathbf{H}^d_{t-1})),\\
\mathbf{C}^d_t &= f^d_t\circ\mathbf{C}^d_{t-1} + i^d_t\circ \tilde{C^d_t}.\\
\end{aligned}
\end{eqnarray}
\newline
Finally, the hidden state $\mathbf{H}^d_t$ and output $\mathbf{X}^d_t$ are updated according to the output state $o_t$ and cell state $\mathbf{C}^d_t$:
\begin{eqnarray}\label{eqn:convLSTM}
\begin{aligned}
o^d_t &= \sigma(W_{d,xo}(\mathbf{X}^{d-1}_{t}) + W_{d,ho}( \mathbf{H}^d_{t-1})),\\
\mathbf{H}^d_t &= o^d_t\circ tanh(\mathbf{C}^d_t), \\
\mathbf{X}^d_t &= \mathbf{H}^d_t,
\end{aligned}
\end{eqnarray}
\noindent where `$\circ$' denotes the Hadamard product. Also, we apply batch normalization \cite{ioffe2015batch} to each convolutional operation.

For every iteration, loss is connected to the output of the last ConvLSTM module in physical depth. Here, the post processes of ConvLSTM module's output (pooling, fully connected layer, etc.) are ignored for sake of simplicity. $\mathbf{L_{t}}$ is the cross entropy loss at time $t$, while $C$ denotes the correct target class number and $\mathbf{L}$ is the overall loss: 
\begin{eqnarray}\label{eqn:loss_connect_to_each}
\begin{aligned}
\mathbf{L} &= \sum_{t=1}^{T} \gamma^t \mathbf{L_{t}} ,  \text{ where }  \mathbf{L_{t}} = -log \frac{e^{\mathbf{H^D_t}[C]}}{\sum_j e^{\mathbf{H^D_t}[j]}} . \\
\end{aligned}
\label{eq:loss}
\end{eqnarray}
$\gamma$ is a constant discount factor determining the worth of early vs later predictions; we set $\gamma=1$ in our experiments which gives equal worth to all iterations.
\footnote{
\textbf{Predicting the `absolute output' vs an `adjustment' value}:
In this formulation, the absolute output is predicted at each iteration. An alternative would be to predict an `adjustment value' at each iteration that, when summed with previous iteration's output, would yield the updated absolute output. This approach would have the disadvantage of being applicable to only output spaces with a numerical structure, e.g. regression problems. Problems without a numerical, e.g. classification, or structured space cannot not be solved using this approach.
}

Connecting the loss to all iterations forces the network to attempt the entire task at each iteration and pass the output via the proxy of hidden state (Eq.~\ref{eq:loss}) to future iterations. Thus, the network cannot adopt a representation scheme like feedforward networks that go from low-level (e.g. edges) to high-level representations as merely low-level representations would not be sufficient for accomplishing the whole classification task in early iterations. Instead, the network forms a representation across iterations in a coarse-to-fine manner (further discussed in sections~\ref{sec:results_early}, ~\ref{sec:taxonomy_results}, and {\href{http://feedbacknet.stanford.edu/supplementary_material}{supplementary material}}'s Sec.~3).

We initialize all $\mathbf{X^0_t}$ as the inout image $inp$, and all $\mathbf{H^d_0}$ as 0, i.e. 
$\forall t\in \{1,2,\cdots, T\}: \mathbf{X}^0_t := inp$ and 
$\forall d\in \{1,2,\cdots, D\}: \mathbf{H}^d_0 := 0$.
The operation of the ConvLSTM module above can be referred to using the simplified notation $\mathfrak{F}(\mathbf{X}^{d-1}_t, \mathbf{H}^d_{t-1})$. 


\subsection{Feedback Module Length} \label{feedback_type_section}
\begin{figure}
  \centering
  \centerline{\includegraphics[width=1\columnwidth, height=6cm]{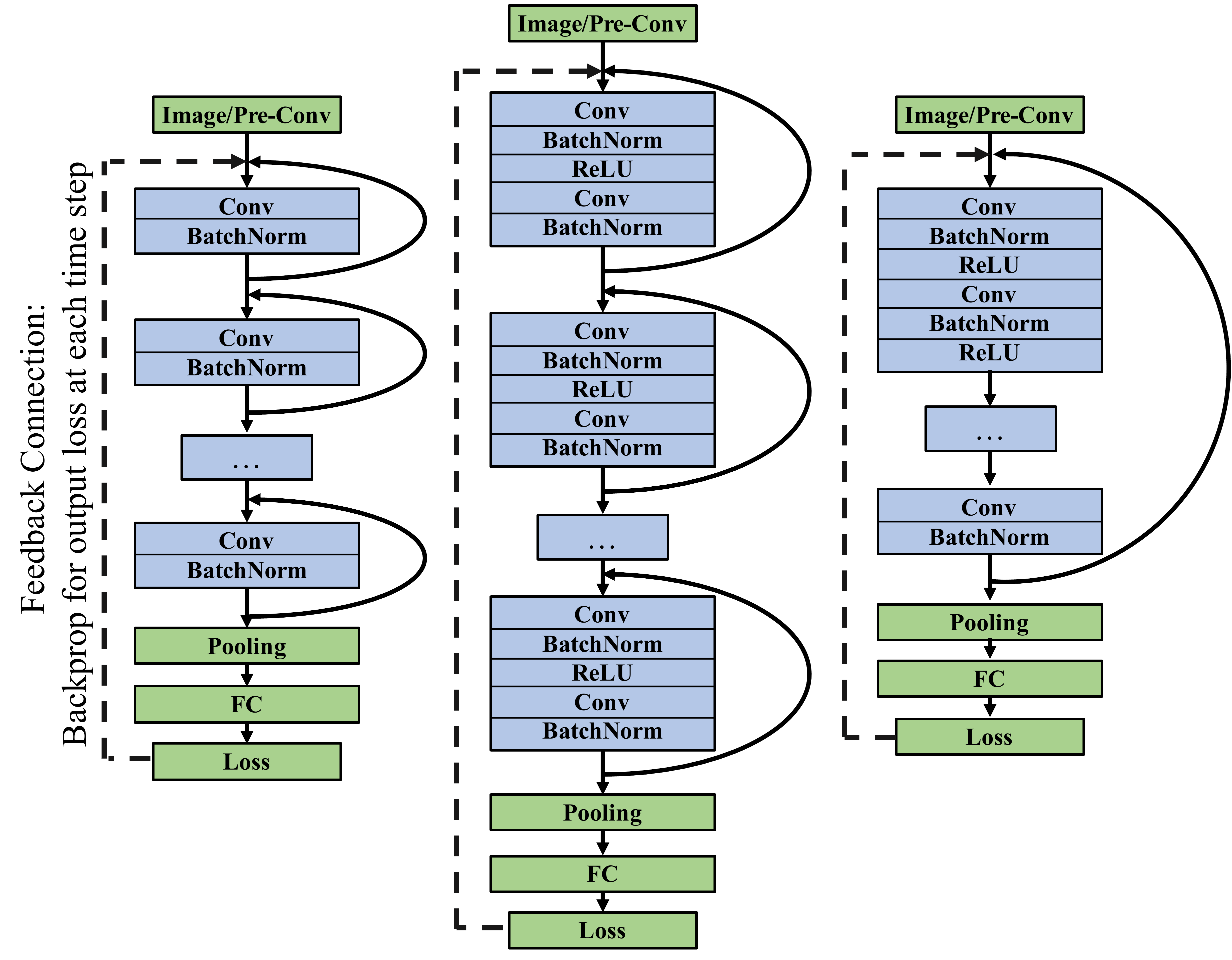}}
  \caption{\footnotesize \textbf{Feedback networks with different feedback module (ConvLSTM) lengths}. Left, middle, and right show Stack-1, Stack-2, and Stack-All, respectively.} \label{fig:feedtype}
\end{figure}

We can stack multiple ConvLSTM modules, each a different number of feedforward layers.
We categorize feedback networks according to the number of feedforward layers (Conv + BN) within one ConvLSTM module, i.e. the local length of feedback. 
This is shown in Fig.~\ref{fig:feedtype} where the models are named Stack-1, Stack-2, and Stack-All. For Stack-$i$, $i$ feedforward layers are stacked within one ConvLSTM module. This essentially determines how distributed the propagation of hidden state throughout the network should be (e.g. for the physical depth $\mathcal{D}$, Stack-All architecture would have one hidden state while Stack-1 would have $\mathcal{D}$ hidden states). See {\href{http://feedbacknet.stanford.edu/supplementary_material}{supplementary material}} (Sec.~2) for more discussions.
Which length $i$ to pick is a design choice; we provide an empirical study on this in Sec.~\ref{sec:feed-type}.

\subsection{Temporal Skip Connection} \label{sec:length_results}
In order to regulate the flow of signal through the network, we include identity skip connections.
This was inspired by conceptually similar mechanisms, such as the residual connection of ResNet~\cite{he2015deep} and the recurrent skip coefficients in~\cite{zhang2016architectural}. The skip connections adopted in the feedback model can be formulated as: with the new input at time $t$ being $\hat{\mathbf{X}^d_t} = \mathbf{X}^d_t + \mathbf{H}^d_{t-n}$, the final representation will be $\mathfrak{F}(\hat{\mathbf{X}^d_t}, \mathbf{H}^d_{t-n}, \mathbf{H}^d_{t-1})$, where $n$ is the skip length. The skip connections are shown in Fig.\ref{fig:skipConnect} denoted by the red dashed lines. We set $n=2$ in our experiments. 

Besides regulating the flow, Table \ref{table:skipconnperf} quantifies the endpoint performance improvement made by such skip connections on CIFAR100~\cite{krizhevsky2009learning} using Stack-2 architecture with physical depth 4 and 8 iterations.

\begin{table}[!h]
\centering
\begin{tabular}{ | c | c | c | }
  \hline      
  Feedback Net              & Top1           & Top5 \\ \hline
  w/o skip connections           & 67.37          & 89.97 \\
  w/ skip connections          & \textbf{67.83} & \textbf{90.12} \\
  \hline
\end{tabular}
\caption{\footnotesize \textbf{Impact of skip connections in time on CIFAR100} ~\cite{krizhevsky2009learning}}
\label{table:skipconnperf}
\end{table}

\newcommand{\cutsectionupTBC}{\vspace*{-0.08in}}
\newcommand{\cutsectiondownTBC}{\vspace*{-0.08in}}
\cutsectionupTBC
\subsection{Taxonomic Prediction}
\label{tax-based-prediction}
\cutsectiondownTBC

It is of particular practical value if the predictions of a model conform to a taxonomy. That is, making a correct coarse prediction about a query, if a correct fine prediction cannot be made. Given a taxonomy on the labels (e.g. ImageNet or CIFAR100 taxonomies), we can examine a network's capacity in making taxonomic predictions based on the fine class's Softmax distribution. 
The probability of a query belonging to the fine class $y_i$ is defined in Softmax as $P(y_i | x; W) = \frac{e^{f_{y_i}}}{\sum_j e^{f_j}}$ for a network with weights $W$. The probability of a query belonging to the $k^{th}$ higher level coarse class $Y_k$ consisting of $\{y_1,y_2,...,y_n\}$ is thus the sum of probability of the query being in each of the fine classes:
\begin{eqnarray}\label{eqn:softmax_coarse}
\begin{aligned}
P(Y_k | x; W) = \sum_{i\in 1:n} P(y_i | x; W) = \frac{\sum_{i\in 1:n} e^{f_{y_i}}}{\sum_j e^{f_j}}.
\end{aligned}
\end{eqnarray}
\noindent Therefore, we use a mapping matrix $M$, where $M(i,k)=1$ if $y_i \in Y_k$, to transform fine class distribution to coarse class distribution. This also gives us the loss for coarse prediction $L^{Coarse}$, and thus, a coarse prediction $p_c$ is obtained through the fine prediction $p_f$. In Sec.~\ref{sec:taxonomy_results}, it will be shown that the outputs of the feedback network conform to a taxonomy especially in early predictions. 

\subsection{Episodic Curriculum Learning}
\label{sec:CL}
As discussed in Sec.~\ref{sec:intro}, the feedback network provides a new way for enforcing a curriculum in learning and enables using a taxonomy as a curriculum strategy. We adopt an iteration-varying loss to enforce the curriculum. We use an annealed loss function at each time step of our $k$-iteration feedback network, where the relationship of coarse class losses $L^{Coarse}_{t}$ and fine class losses $L^{Fine}_{t}$ parametrized by time $t$ is formulated as:
\begin{equation}\label{eq:2}
\begin{aligned}
L(t) = \zeta L^{Coarset}_{t} +  (1 - \zeta) L^{Fine}_{t},\\
\end{aligned}
\end{equation}
\noindent where $\zeta$ is the weights that balance the contribution of coarse and fine losses. We adopt a linear decay as $\zeta = \frac{t}{k}$, where $t = 0, 1,..., k$, and $k$ is the end iteration of decaying.

For object classification, the time varying loss function encourages the network to recognize objects in a first coarse then fine manner, i.e. the network learns from the root of an taxonomy tree to its leaves. In Sec.~\ref{sec:results_CL}, it will be empirically shown that the feedback based approach well utilizes this curriculum strategy.


\subsection{Computation Graph Analysis} 
\label{complexity_analysis}
\begin{figure}
  \centering
  \centerline{\includegraphics[width=1\columnwidth]{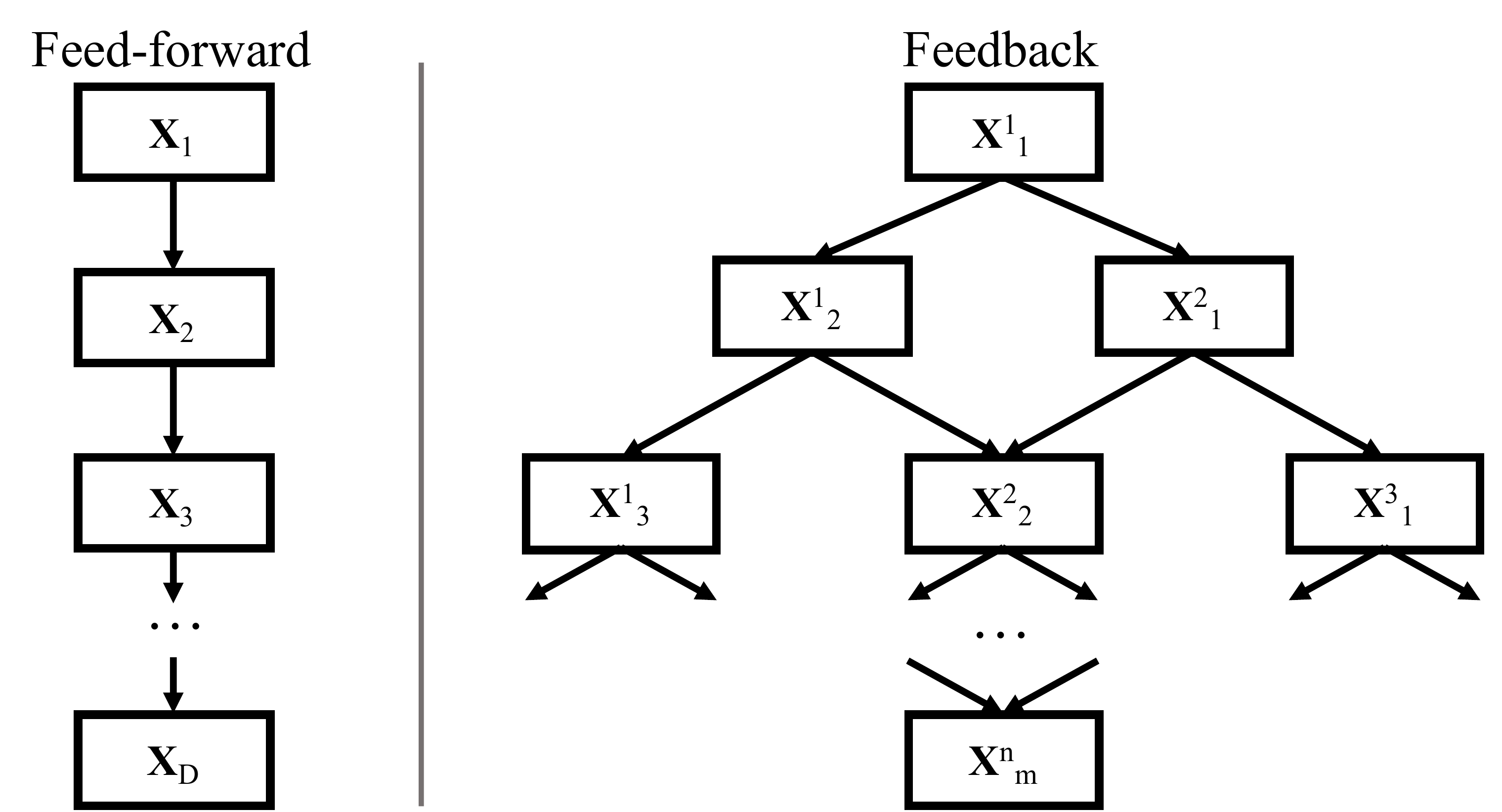}}
  \caption{\footnotesize \textbf{Computation graph of Feedback vs Feedforward.} $\mathbf{X}^j_{i}$ denotes the representation at temporal iteration $i$ and physical depth $j$. Skip connections are not shown for simplicity.} 
  \label{fig:computationGraph}
\end{figure}
Under proper hardware, feedback model also has an advantage on speed over feedforward. This is because a feedback network is a better fit for parallelism compared to feedforward due to having a shallower computation graph (shown in Fig.~\ref{fig:computationGraph}). In the interest of space, we give the full discussion and derivation of the computation graphs in {\href{http://feedbacknet.stanford.edu/supplementary_material}{supplementary material}} (Sec.~4) and only compare their depths here. The computation graph depth of feedforward model with depth $\mathcal{D}$ and that of feedback model with same virtual depth (consisting of $m$ temporal iterations and physical depth $n$, $\mathcal{D}=m\times n$, and Stack-1 configuration) are $d_{ff}= \mathcal{D}-1=mn-1$ and $d_{fb}=m+n-1$, respectilvey.

 Under a proper hardware scenario where one can do parallel computations to a sufficient extent, inference time can be well measured by the longest distance from root to target (i.e. graph's depth). Therefore, the total prediction time of feedforward network is larger than feedback network's as $d_{ff} = mn - 1 > m + n-1=d_{fb}$. Please see {\href{http://feedbacknet.stanford.edu/supplementary_material}{supplementary material}} (Sec.~4) for the depth comparison for early predictions, Stack-$i$ configuration, and traning time.


\begin{figure*}[t!h]
\begin{multicols}{2}
    \includegraphics[width=\linewidth]{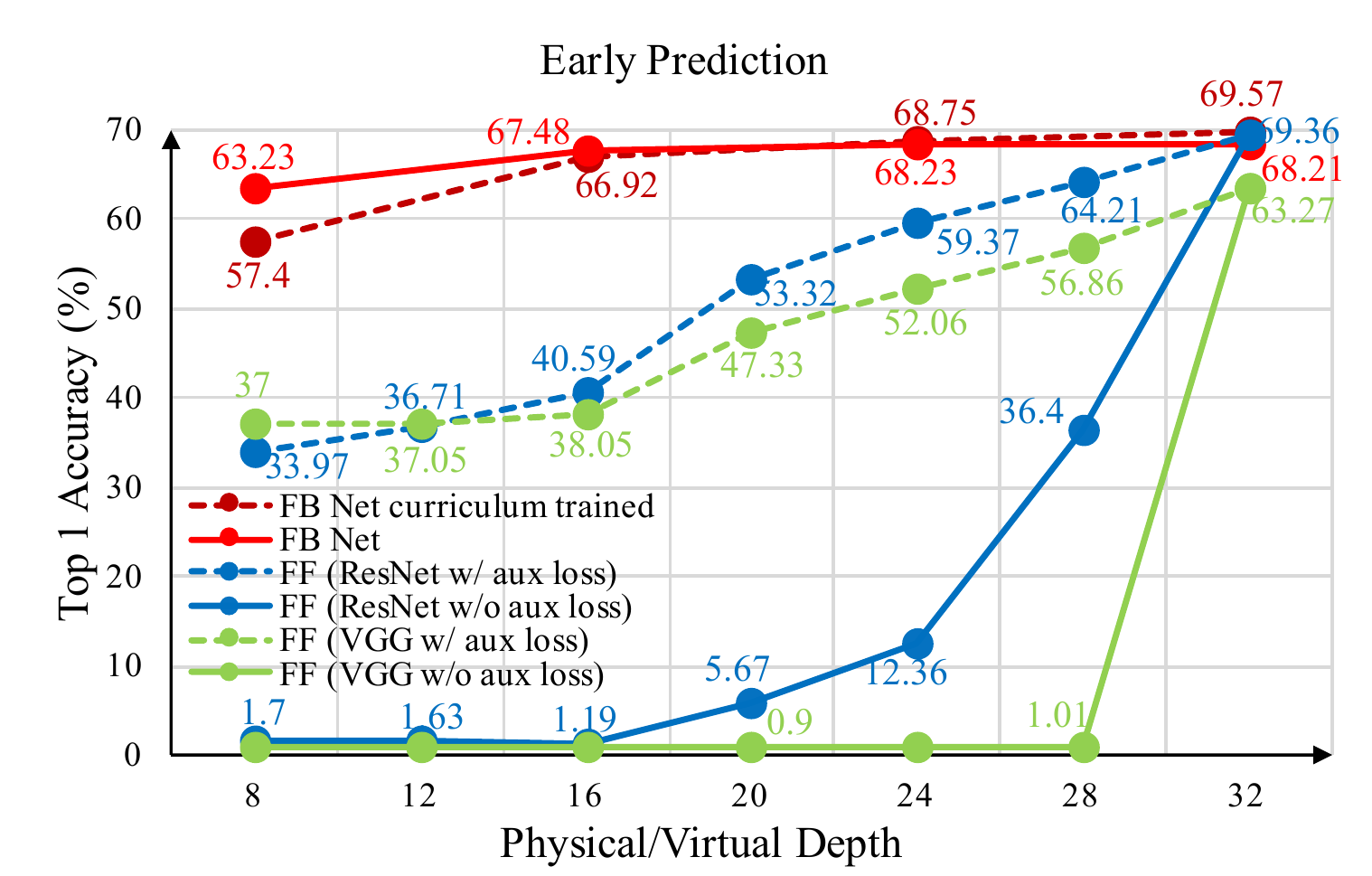}\par 
    \caption{\footnotesize \textbf{Evaluation of early predictions.} Comparison of accuracy of feedback (FB) model and feedforward (FF) baselines (ResNet \& VGG, with or without auxiliary loss layers)} \label{fig:early}
    \includegraphics[width=\linewidth]{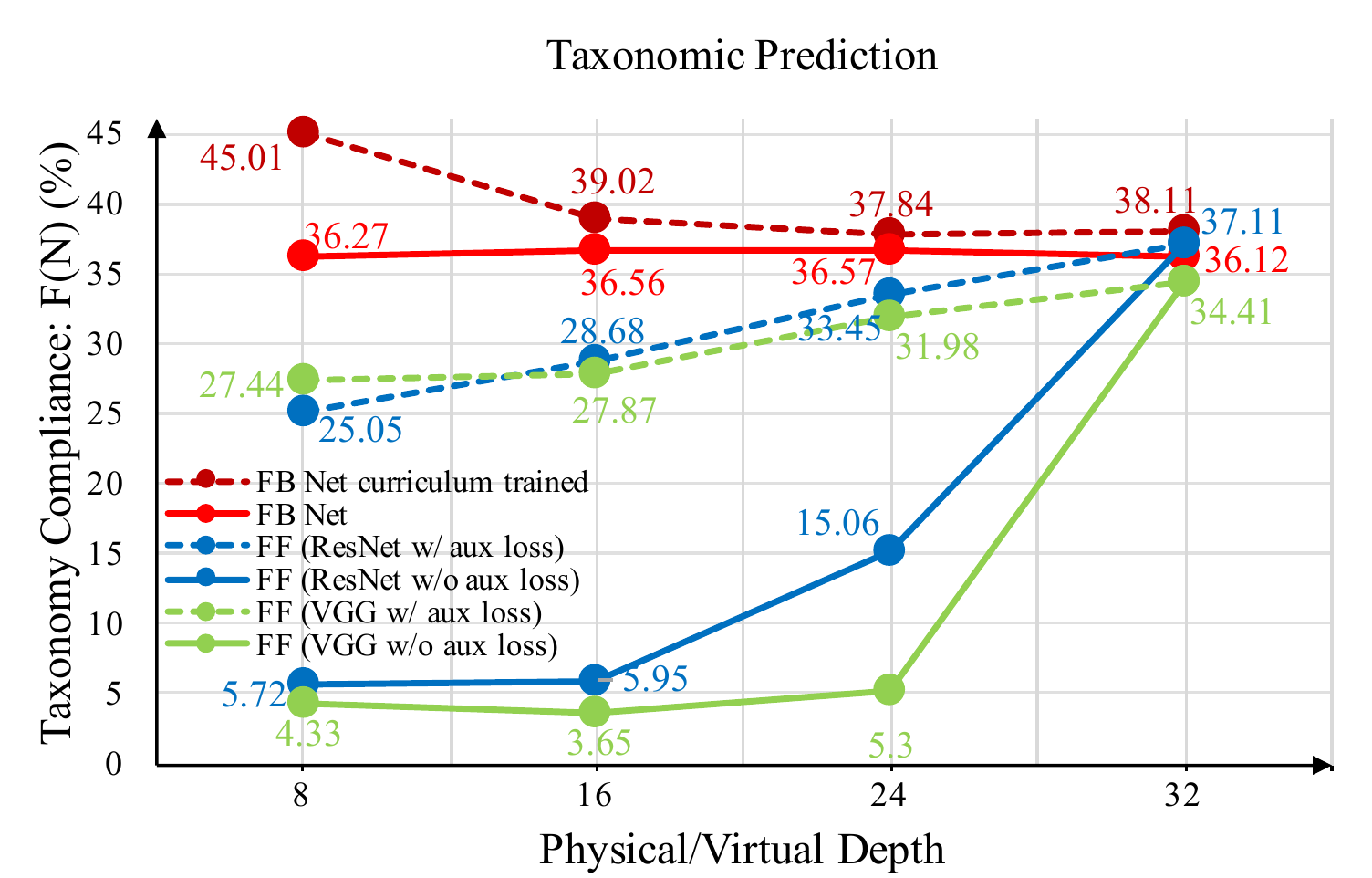}\par
    \caption{\footnotesize \textbf{Evalaution of taxonomoy based prediction for feedback (FB) and feedforward (FF) networks trained with or without auxiliary layers.} We use only fine loss to train, except for the curriculum learned one.}
    \label{cfloss}
\end{multicols}
\end{figure*}
\section{Experimental Results}
Our experimental evaluations performed on the three benchmarks of CIFAR100~\cite{krizhevsky2009learning}, Stanford Cars~\cite{krause20133d}, and MPII Human Pose~\cite{andriluka20142d}, are provided in this section. 
\subsection{Baselines and Terminology} 
\label{baseline_def}
Below we define our terminology and baselines: 
\newline
\noindent\textbf{Physical Depth}: the depth of the convolutional layers from input layer to output layer. For feedback networks, this represents the number of stacked physical layers across all ConvLSTM modules ignoring the temporal dimension.
\newline
\noindent\textbf{Virtual Depth}: physical depth \(\times\) number of iterations. This is the effective depth considering both spatial and temporal dimensions. (not applicable to feedforward models.)
\newline
\noindent\textbf{Baseline Models}: We compare with ResNet\cite{he2015deep} and VGG\cite{simonyan2014very} as two of the most commonly used feedforward models and with closest architecture to our convolutional layers. Both baselines have the same architecture, except for the residual connection.
We use the same physical module architecture for our method and the baselines. We also compare with ResNet original authors' architecture~\cite{he2015deep}. The kernel sizes and transitions of filter numbers remain the same as original paper's. In Sec.~\ref{sec:MPIPose}, we compare with feedforward Hourglass~\cite{newell2016stacked} by making a feedback Hourglass.
\newline
\noindent\textbf{Auxiliary prediction layer (aux loss)}: Feedfoward baselines do not make episodic or mid-network predictions. In order to have a feedforward based baseline for such predictions, we train new pooling$\rightarrow$FC$\rightarrow$loss layers for different depths of the feedforward baselines (one dedicated aux layers for each desired depth). This allows us to make predictions using the mid-network representations. We train these aux layers by taking the fully trained feedforward network and training the aux layers from shallowest to deepest layer while freezing the convolutional weights.


\subsection{CIFAR-100 and Analysis}
CIFAR100 includes 100 classes containing 600 images each. The 100 classes (fine level) are categorized into 20 classes (coarse level), forming a 2-level taxonomy. All of the reported quantitative and qualitative results were generated using the fine-only loss (i.e. the typical 100-way classification of CIFAR100), unless specifically mentioned curriculum learning or coarse+fine loss (Eq.~\ref{eq:2}) were used. 


\cutsectionup
\subsubsection{Feedback Module Length}
\label{sec:feed-type}
\cutsectiondown
Table~\ref{table:feedbackTypes} provides the results of feedback module length study per the discussion in Sec.~\ref{feedback_type_section}. The physical depth and iteration count are kept constant (physical depth 4 and 4 iterations) for all models.
The best performance is achieved when the local feedback length is neither too short nor too long. We found this observation to be valid across different tests and architectures, though the optimal length may not always be 2.  In the rest of the experiments for different physical depths, we optimize the value of this hyperparameter empirically (often ends up as 2 or 3). See  {\href{http://feedbacknet.stanford.edu/supplementary_material}{supplementary material}}'s Sec.~6 for an experimental discussions on the trade-off between physical depth and iteration count as well as optimal iteration number.

\begin{table}[!htb]
\centering
\begin{tabular}{ | c | c | c | }
  \hline      
  Feedback Type & Top1           & Top5  \\ \hline
  Stack-1       & 66.29          & 89.58 \\
  Stack-2       & \textbf{67.83} & \textbf{90.12} \\
  Stack-All     & 65.85          & 89.04 \\
  \hline
\end{tabular}
\caption{\footnotesize \textbf{Comparison of different feedback module lengths,} all models have the same physical depth 4 and virtual depth 16.}
\label{table:feedbackTypes}
\end{table}

\newcommand{\cutsectionupEP}{\vspace*{-0.15in}}
\newcommand{\cutsectiondownEP}{\vspace*{-0.08in}}
\cutsectionupEP


\subsubsection{Early Prediction} \label{sec:results_early}
\cutsectiondownEP

We evaluate early predictions of various networks in this section. We conduct this study using a feedback network with virtual depth 32 (similar trends achieved with other depths) and compare it with various feedforward networks. As shown in Fig.~\ref{fig:early}, at virtual depths of 8, 12, and 16, the feedback network already achieves satisfactory and increasing accuracies. The solid blue and green curves denote the basic feedforward networks with 32 layers; their rightmost performance is their endpoint results, while their early predictions are made using their final pooling$\rightarrow$FC$\rightarrow$loss layer but applied on mid-network representations. The dashed blue and green curves show the same, with the difference that the trained pooling$\rightarrow$FC$\rightarrow$loss layers (aux loss, described in Sec.~\ref{baseline_def}) are employed for making early predictions. The plot shows that the feedforward networks perform poorly when using their first few layers' representations, confirming that the features learned there are not suitable for completing the ultimate output task (expected)~\cite{zeiler2014visualizing}. This is aligned with the hypothesis that feedback model forms its representation in a different and coarse-to-fine manner (further discussed in Sec.~\ref{sec:taxonomy_results}).

We also attempted full training and fine tuning the feedforward networks with aux losses, but this never led to a better performance than the reported curves in Fig.~\ref{fig:early} by sacrificing either early or endpoint performances. The best results were  (comparable to curves in Fig.~\ref{fig:early}): 6.8\%, 10.2\%, 13.1\%, 13.0\%, 59.8\%, 66.3\%, 68.5\% for depths 8, 12, 16, 20, 24, 28, and 32, respectively.

\noindent\textbf{Comparison with Feedforward Ensemble:} Although it is memory inefficient and wasteful in training, one can also achieve the effect of early prediction through an ensemble of feedforward models in parallel (i.e. for every depth at which one desires a prediction, have a dedicated feedforward network of that depth). Since running an ensemble of ResNets in parallel has similar optimal hardware requirements of Sec. \ref{complexity_analysis}, we make a comparison under the same analysis: applying the computation graph depth analysis to a 48 layer virtual depth feedback model (physical depth 12, Stack-3, 4 iterations), if we denote the time to finish one layer of convolution as $T$, then we have $i^{th}$ iteration result at: $t_i=(12+3i) T$. Then the first to last iterations' results (virtual depths $12,24,36,48$) will become available at $12T,15T,18T,$ and $21T$. To have ResNet results at the same times, we need an ensemble of ResNets with depths $12,15,18,21$. The performance comparison between feedback network and the ensemble is provided in Table \ref{table:early_ensemble}, showing the advantage of feedback networks.
\begin{table}[h]
\centering
\begin{tabular}{  c | c | c | c | c |}
  \cline{2-5} & \multicolumn{4}{ |c| }{Time Steps} \\ \hline
  \multicolumn{1}{|c|}{Model} & 12T & 15T & 18T & 21T  \\ \hline
  \multicolumn{1}{|c|}{Feedback Network} & \textbf{67.94} &\textbf{70.57} &\textbf{71.09} &\textbf{71.12} \\ \hline
  \multicolumn{1}{|c|}{ResNet Ensemble} & 66.35 &67.52 &67.87 & 68.2 \\
  \hline
\end{tabular}
\caption{\footnotesize \textbf{Top1 accuracy comparison between Feedback Net and an ensemble of ResNets} that produce early predictions at the same computation graph depth time steps.}
\label{table:early_ensemble}
\end{table}

\noindent\textbf{Feedback vs No Feedback:} To examine whether the offered observations are caused by feedback or only the recurrence mechanism, we performed a test by disconnecting the loss from all iterations except the last, thus making the model recurrent feedforward. As shown in Table \ref{table:sequencer_effect}, making the model recurrent feedforward takes away the ability to make early and taxonomic predictions (discussed next).
\begin{table}[htb]
\centering
\begin{tabular}{  c | c | c | c | c |}
  \cline{2-5} & \multicolumn{4}{ |c| }{Virtual Depth} \\ \hline
   \multicolumn{1}{|c|}{Model} & 12 & 24 & 36 & 48  \\ \hline
  \multicolumn{1}{|c|}{\footnotesize{Feedback}}  & \textbf{67.94} &\textbf{70.57} &\textbf{71.09} &71.12 \\ \hline
   \multicolumn{1}{|c|}{\footnotesize{Feedback Disconnected}} & \multirow{2}{*}{36.23} &\multirow{2}{*}{62.14} &\multirow{2}{*}{67.99 }&\multirow{2}{*}{\textbf{71.34}}  \\
   \multicolumn{1}{|c|}{\footnotesize{(Recurrent Feedforward)}} &   &  &  &  \\   
  \hline
\end{tabular}
\caption{\footnotesize \textbf{The impact of feedback} on CIFAR100 for a model with virtual depth 48 and four iterations.}
\label{table:sequencer_effect}
\end{table}

\newcommand{\cutsectionupEPC}{\vspace*{-0.15in}}
\newcommand{\cutsectiondownEPC}{\vspace*{-0.08in}}
\cutsectiondownEPC


\subsubsection{Taxonomic Prediction} 
\label{sec:taxonomy_results}

\begin{figure}
\centerline{\includegraphics[width=1\columnwidth]{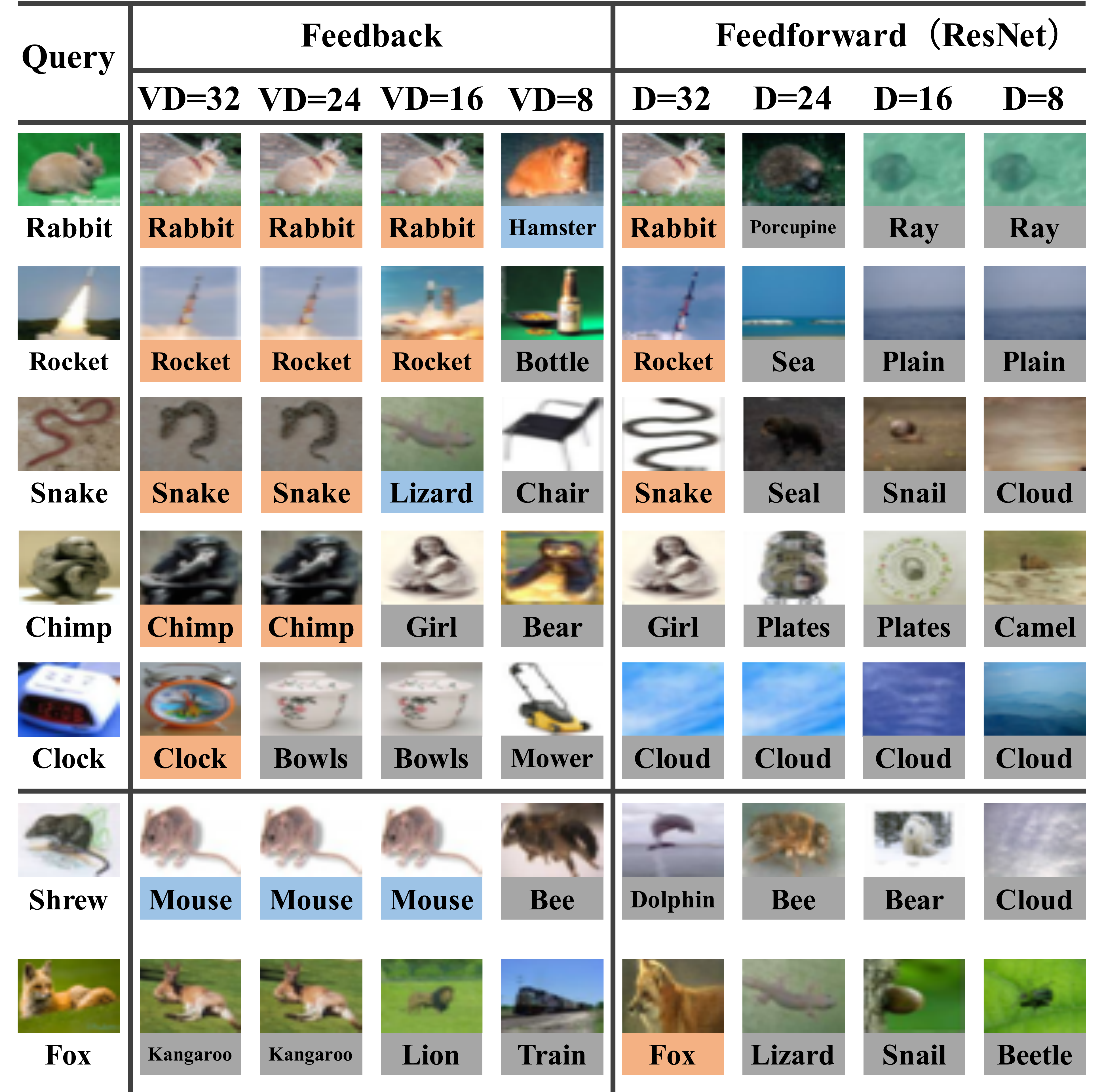}}
\caption{\footnotesize \textbf{Qualitative results of classification on CIFAR100.} Each row shows a query along with nearest neighbors at different depths for feedback and feedfowrad networks. Orange, blue, and gray represent `correct fine class', `correct coarse class but wrong fine class', and `both incorrect', respectively. Two bottom queries are representative failure cases.}
\label{fig:nnfigure}
\end{figure}

We measure the capacity $F({N})$ of network ${N}$ in making taxonomic predictions (taxonomy compliance) as: the probability of making a correct coarse prediction for a query if it made a wrong fine prediction for it; in other words, how effective it can correct its wrong fine class prediction to a correct coarse class: $F({N})=P(correct(p_c)|!correct(p_f); {N}).$ As defined in Sec.~\ref{tax-based-prediction}, $pc$ and  $pf$ stand for coarse and fine prediction, respectively.

\begin{figure}
  \centering
  \centerline{\includegraphics[width=1\columnwidth]{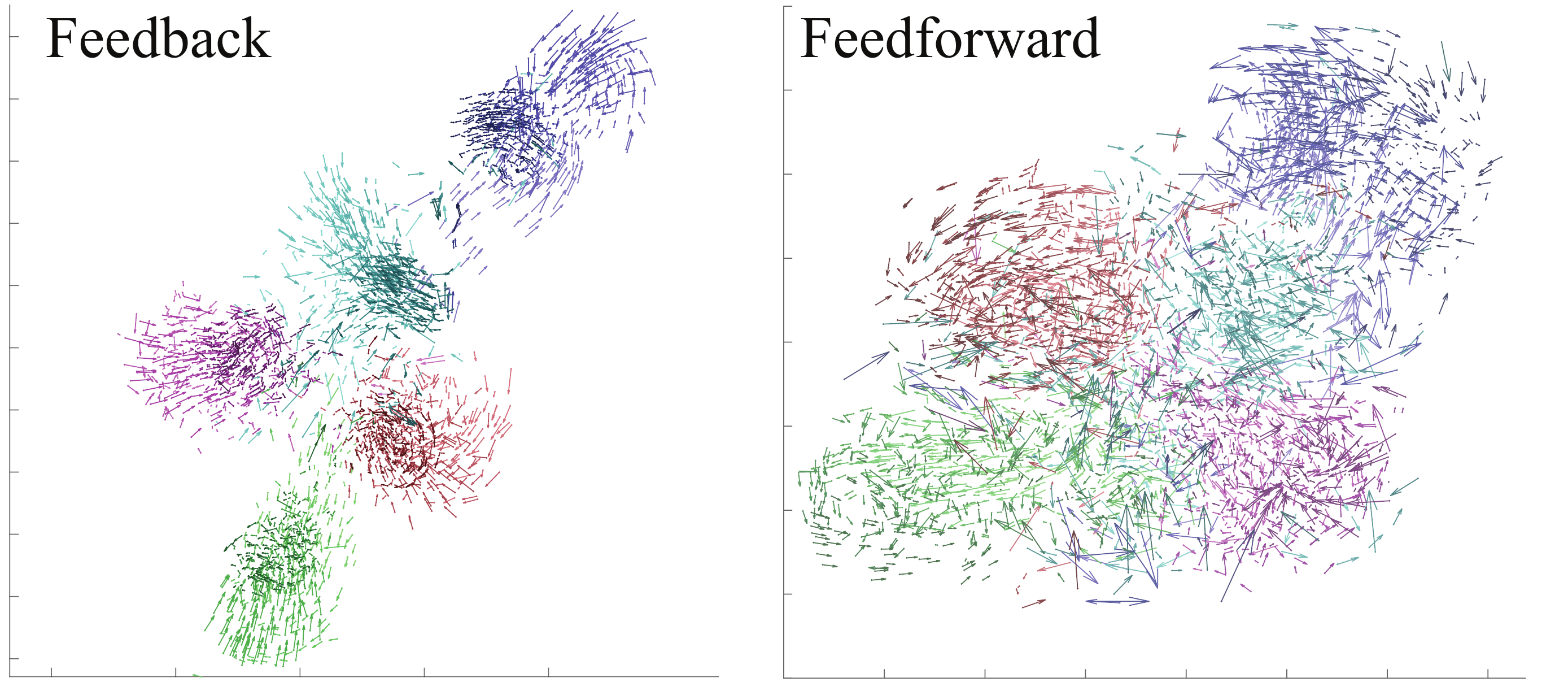}}
  \caption{\footnotesize \textbf{Timed-tSNE plots showing how the representation evolves through depth/iterations (i.e. how a datapoint moved in representation space)} for each method, on five random classes of CIFAR100. The lighter the hue of the arrow, the earlier the depth/iteration. Feedback's representation is relatively disentangled throughout, while feedforward's representation gets disentangled only towards the end. (Best see on screen. Vector lengths are shown in half to avoid cluttering.)} 
  \label{fig:timedtsne}
\end{figure}

The quantitative and qualitative results are provided in Figures~\ref{cfloss}, \ref{fig:nnfigure}, and~\ref{fig:timedtsne}. Note that all of these results were naturally achieved, i.e. using fine-only loss and no taxonomy or curriculum learning was used during training (except for the dashed red curve which was trained using curriculum learning; Sec.~\ref{sec:results_CL}). Fig.~\ref{cfloss} shows feedback network's predictions better complies with a taxonomy even at shallow virtual depths, while feedforward model does not achieve the same performance till the last layer, even when using dedicated auxiliary layers. This is again aligned with the hypothesis that the feedback based approach develops a coarse-to-fine representation and is observed in both figures~\ref{fig:nnfigure} and \ref{fig:timedtsne}. In Fig.~\ref{fig:nnfigure}, early prediction classes and nearest neighbor images (using the network representations) for both feedback and feedforward networks are provided, showing significantly more relevant and interpretable early results for feedback.  

\noindent\textbf{Timed-tSNE:} In Fig.~\ref{fig:timedtsne}, we provide a variant of tSNE~\cite{maaten2008visualizing} plot which we call \emph{timed-tSNE}. It illustrates \emph{how the representation of a network evolves} throughout depth/iterations, when viewed through the window of class labels. For each datapoint, we form a temporally regulated trajectory by connecting a set of 2D tSNE embedding locations. For feedback network, the embeddings of one datapoint come from the representation at different iterations (i.e. $i$ embeddings for a network with $i$ iterations). For feedforward, embeddings come from difference layers. More details provided in {\href{http://feedbacknet.stanford.edu/supplementary_material}{supplementary material}} (Sec.~5).

Fig.~\ref{fig:timedtsne} suggests that feedforward representation is intertwined at early layers and disentangles the classes only in the last few layers, while feedback's representation is disentangled early on and the updates are mostly around forming fine separation regions. This again supports the hypothesis that feedback develops a coarse-to-fine representation. We also provide activation maps of feedback vs feedforward models in {\href{http://feedbacknet.stanford.edu/supplementary_material}{supplementary material}} (Sec.~5.2) exhibiting notably dissimilar patterns, and thus, dissimilar representations, thought their endpoint numerical results are close.


\vspace{-10pt}
\subsubsection{Curriculum Learning}
 \label{sec:results_CL}

Table~\ref{table:coarsefine} compares the performance of the networks when trained with the fine-only loss vs the episodic coarse-to-fine curriculum loss (Sec.~\ref{sec:CL}). We employed the same episodic curriculum training for the feedback network and the baselines ``w/ Aux loss'', while the baselines ``w/o Aux loss" had to use conventional curriculum training (datapoint sorting)~\cite{bengio2009curriculum}.
The best performance with highest boost is achieved by feedback network when using curriculum learning. Also, using the episodic curriculum training improves taxonomic prediction results as shown by the curriculum curve in Fig.~\ref{cfloss}.

\begin{table}[htb]
\centering
\begin{tabular}{|c| p{0.3cm} | l | l |}
  \hline      
   Model                 &CL    &  Top1(\%)-\footnotesize{Fine}  &  Top1(\%)-\footnotesize{Coarse} \\ \hline
  Feedback Net          & N         &  68.21                & 79.7 \\ \cline{2-4}
                        & Y         &  \textbf{69.57}\footnotesize{(+1.34\%)}       & \textbf{80.81}\footnotesize{(+1.11\%)} \\ \hline
  Feedforward           & N         &  69.36                & 80.29          \\ \cline{2-4}
  \footnotesize ResNet w/ Aux loss  & Y         &  69.24\footnotesize{(-0.12\%)}               & 80.20\footnotesize{(-0.09\%)}           \\ \hline
  Feedforward           & N        &  69.36               & 80.29          \\ \cline{2-4}
  \footnotesize ResNet w/o Aux loss & Y         &  65.69\footnotesize{(-3.67\%)}               & 76.94\footnotesize{(-3.35\%)}          \\ \hline
  Feedforward           & N         & 63.56            & 75.32       \\ \cline{2-4}
  \footnotesize VGG w/ Aux loss  & Y         & 64.62\footnotesize{(+1.06\%)}               & 77.18\footnotesize{(+1.86\%)}           \\ \hline
  Feedforward           & N         & 63.56                & 75.32     \\ \cline{2-4}
  \footnotesize VGG w/o Aux loss & Y         & 63.2\footnotesize{(-0.36\%)}               & 74.97\footnotesize{(-0.35\%)}          \\ \hline
\end{tabular}
\caption{\footnotesize \textbf{Evaluation of the impact of Curriculum Learning  (CL)} on CIFAR100. The CL column denotes if curriculum learning was used. The difference made by curriculum for each method is shown in parentheses.}
\label{table:coarsefine}
\end{table}


\subsubsection{Endpoint Performance Comparison} 
\label{end-to-end-perf}
\cutsectiondownEPC
Table~\ref{table:comparisons} compares the endpoint performance of various feedforward and feedback models on CIFAR100. The detailed architecture of each model is provided in the end of this section. Feedback networks outperform the baselines with the same physical depth by a large margin and work better than or on part with baselines with the same virtual depth or deeper. This ensures that the discussed advantages in early and taxonomic prediction were not achieved at the expense of sacrificing the endpoint performance. 

The bottom part of Table~\ref{table:comparisons} shows several recent methods that are not comparable to ours, as they employ additional mechanisms (e.g. stochasticity in depth~\cite{huang2016deep}) which we did not implement in our model. Such mechanisms are independent of feedback and could be used concurrently with it, in the future. However, we include them for the sake of completeness.

\noindent\textbf{Architectures:} The detailed architectures of feedback and feedforward networks are:~\footnote{The following naming convention is used: $C(fi,fo, k, s)$: $fi$ input and $fo$ output convolutional filters, kernel size $k \times k$, stride $s$. $ReLU$: rectified linear unit. $BN$: batch normalization. $BR=BN+ReLU$. $Avg(k, s)$: average pooling with spatial size $k \times k$, and stride s. $FC(fi,fo)$: fully connected layer with $fi$ inputs, and $fo$ outputs.}
\newline
$\bullet$ Recurrent Block: {\footnotesize $Iterate(fi,fo, k, s, n, t)$} denotes our convLSTM recurrent module (defined in Sec.~\ref{functions_section}) which iterates $t$ times and has gate functions, i.e. $W$, with the feedforward architecture:

{\footnotesize $\rightarrow$ $C(fi,fo, k, s) \rightarrow BR$ $\rightarrow$ $\{C(fo,fo, k, 1)\rightarrow BR\}^{n-1}$. }
\newline
We denote stacking using $\{...\}^n$ indicating that the module in brackets is stacked $n$ times. We use the same architecture as above for all gates and include residual connections in it.
\newline
$\bullet$ Preprocess and Postprocess: across all models, we apply the following pre-process: {\footnotesize $Input$ $\rightarrow$ $C(3,16, 3, 1)\rightarrow BR$} and post-process: {\footnotesize $\rightarrow$ $Avg(8,1)$ $\rightarrow$ $FC(64,100)$}
\newline
$\bullet$ Feedback Network with physical depth = 8: 

{\footnotesize 
$\rightarrow$ $Iterate(16,32, 3, 2, 2, 4)$ $\rightarrow$ $Iterate(32,32, 3, 1, 2, 4)$

$\rightarrow$ $Iterate(32,64, 3, 2, 2, 4)$ $\rightarrow$ $Iterate(64,64, 3, 1, 2, 4)$}
\newline
$\bullet$ Feedback Network with physical depth = 12: 

{\footnotesize
$\rightarrow$ $Iterate(16,16, 3, 1, 3, 4)$ $\rightarrow$ $Iterate(16,32, 3, 2, 3, 4)$

$\rightarrow$ $Iterate(32,64, 3, 2, 3, 4)$ $\rightarrow$ $Iterate(64,64, 3, 1, 3, 4)$}
\newline
$\bullet$ Baseline Feedforward models with physical depth = $\mathcal{D}$: 

{\footnotesize
$\rightarrow$ $C(16,32, 3, 2)\rightarrow BR$ $\rightarrow$ $\{C(32,32, 3, 1) \rightarrow BR\}^{\frac{\mathcal{D}}{2}-1}$

$\rightarrow$ $C(32,64, 3, 2) \rightarrow BR$ $\rightarrow$ $\{C(64,64, 3, 1) \rightarrow BR\}^{\frac{\mathcal{D}}{2}-1}$}

\begin{table}
\centering
\begin{tabular}{ | c | c | c | c | c | }
  \hline      
  Model                     & \footnotesize{Physical}   & \footnotesize{Virtual}  & Top1 & Top5  \\ 
                            &   \footnotesize{Depth}   &  \footnotesize{Depth}   &  (\%) &(\%)  \\ \hline
    \textbf{ Feedback Net}   & 12        & 48       & \textbf{71.12}  &\textbf{91.51} \\
    \smash{\raisebox{-0.6\height}{\includegraphics[height=8mm]{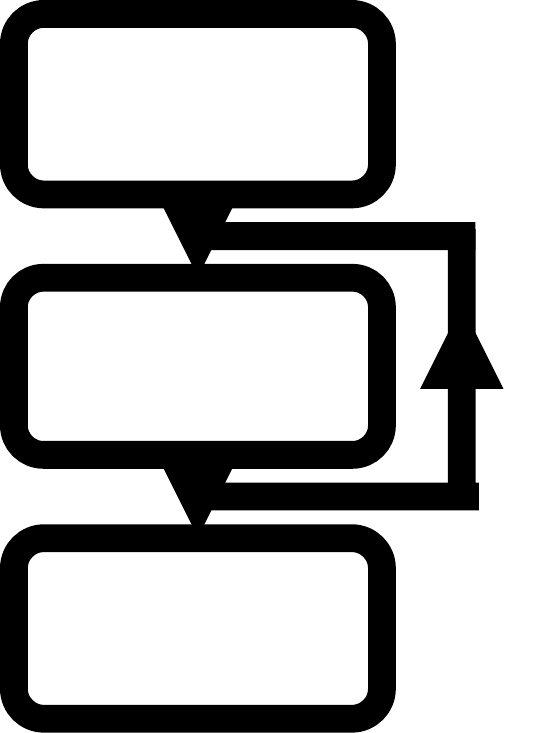}}}
                            &  8        & 32       & 69.57           & 91.01        \\
                            & 4         & 16     & 67.83            &90.12            \\\hline                            
                            & 48        & -        & 70.04           & 90.96        \\
                            & 32        & -        & 69.36           & 91.07        \\
     Feedforward   & 12       & -               & 66.35        & 90.02 \\
    \footnotesize(ResNet\cite{he2015deep})
                            &  8        & -        & 64.23           & 88.95        \\ \cline{2-5}
    \smash{\raisebox{-0.6\height}{\includegraphics[height=8mm]{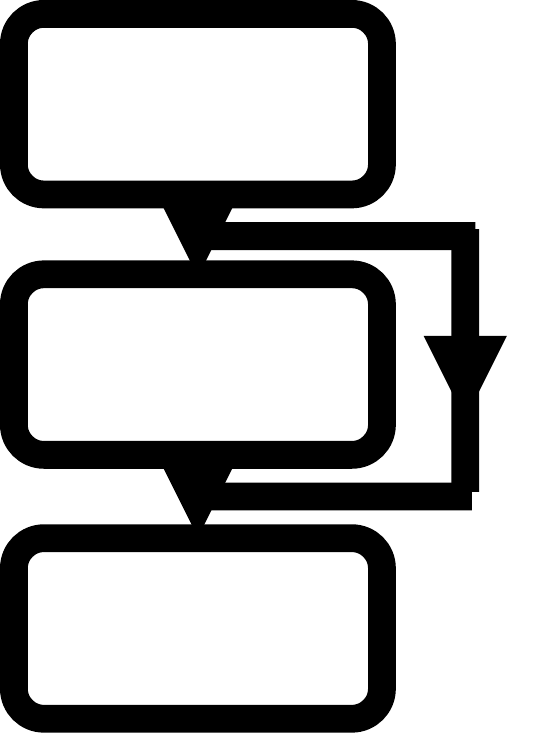}}}         
                            & 128*       & -        & 70.92           & 91.28       \\
                            & 110*       & -        & 72.06           & 92.12       \\
                            & 64*       & -        & 71.01           & 91.48        \\
                            & 48*       & -        & 70.56           & 91.60        \\
                            & 32*       & -        & 69.58           & 91.55        \\ \hline
     Feedforward &48 &- &55.08 &82.1 \\
    \footnotesize(VGG\cite{simonyan2014very})        & 32        & -        & 63.56           & 88.41        \\
    \smash{{\raisebox{-0.6\height}{\includegraphics[height=8mm]{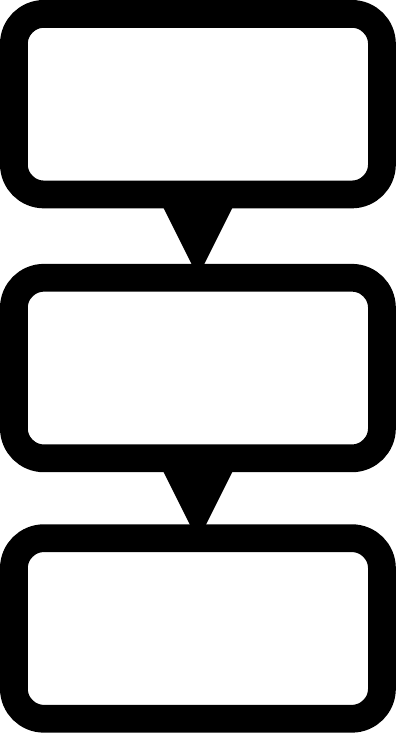}}}}
                            & 12        & -        & 64.65           & 89.26        \\
                            &  8        & -        & 63.91           & 88.90        \\ \hline
                            \hline 
    {Highway \cite{srivastava2015highway}}      &19            &- & 67.76     &-           \\
    {ResNet v2\cite{he2016identity}}      &1001          &-&77.29            &-           \\    
    {\small{Stochastic Depth} \cite{huang2016deep}}      &110             &- &75.02          &-            \\
    {SwapOut \cite{singh2016swapout}}      &32 fat             &- &77.28          &-            \\
    {RCNN \cite{liang2015recurrent}}      &4 fat           &16 &68.25          &-          \\
    \hline
\end{tabular}
\caption{\footnotesize \textbf{Endpoint performance comparison on CIFAR-100.} Baselines denoted with * are the architecture used in the original ResNet paper.}
\label{table:comparisons}
\end{table}


\subsection{Stanford Cars Dataset}

To verify the observations made on CIFAR100 on another dataset, we performed the same set of experiments on Stanford Cars dataset~\cite{krause20133d}.
Evaluations of endpoint performance and curriculum learning are provided in table~\ref{table:cars}. Early prediction and taxonomic prediction curves are provided in {\href{http://feedbacknet.stanford.edu/supplementary_material}{supplementary material}} (Sections 8.1 and 8.2). The experiments show similar trends to CIFAR100's and duplicate the same observations. 

\begin{table}[htb]
\centering
\begin{tabular}{|c| p{0.3cm} | l | l |}
  \hline      
  Model                 &CL         &  Fine     & Coarse \\ \hline
  Feedback Net          & N         &  50.33    & 74.15 \\ \cline{2-4}
                        & Y         &  \textbf{53.37\footnotesize{(+3.04\%)}} 
                                    & \textbf{80.7\footnotesize{(+6.55\%)}} \\ \hline
  Feedforward           & N         &49.09     & 72.60 \\ \cline{2-4}
  \footnotesize{ResNet-24} &Y          &50.86\footnotesize{(+1.77\%)}
                                    &77.25\footnotesize{(+4.65\%)}  \\ \hline
  Feedforward           & N         & 41.04 & 67.65 \\ \cline{2-4}
  \footnotesize{VGG-24}  &Y &41.87\footnotesize{(+0.83\%)}                                                                      &70.23\footnotesize{(+2.58\%)}  \\ \hline
\end{tabular}
\caption{\footnotesize \textbf{Evaluations on Stanford Cars dataset}. The CL column denotes if curriculum learning was employed. All methods have (virtual or physical) depth of 24.}
\label{table:cars}
\end{table}


All networks were trained from scratch without fine-tuning pretrained ImageNet~\cite{deng2009imagenet} models~\cite{lin2015bilinear} or augmenting the dataset with additional images~\cite{xie2015hyper}. To suit the relatively smaller amount of training data in this dataset, we use shallower models for both feedforward and feedback: feedforward baselines have depth of 24 and feedback network has physical depth 6 and iteration count 4, following the same design in Sections~\ref{baseline_def}~\&~\ref{end-to-end-perf}. Full experimental setup is provided in {\href{http://feedbacknet.stanford.edu/supplementary_material}{supplementary material}} (Sec.~8).

\subsection{Human Pose Estimation}
\label{sec:MPIPose}
\vspace{-2pt}
We evaluated on the regression task of MPII Human Pose estimation~\cite{andriluka20142d} benchmark which consists of 40k samples (28k training, 11k testing). Just like we added feedback to feedforward models for CIFAR100 classification and performed comparisons, we applied feedback to the state of the art MPII model Hourglass~\cite{newell2016stacked}. We replaced the sequence of ResNet-like convolutional layers in one stack Hourglass with ConvLSTM, which essentially repalced physical depth with virtual depth, and performed backprobapation at each iteration similar to the discussion in Sec.~\ref{functions_section} (more details about the architecture provided in {\href{http://feedbacknet.stanford.edu/supplementary_material}{supplementary material}}). The performance comparison in Table~\ref{table:mpii} shows that the feedback model outperforms the deeper feedforward baseline. We provide more results and comparisons with other feedback based methods~\cite{carreira2015human,belagiannis2016recurrent} on this benchmark in {\href{http://feedbacknet.stanford.edu/supplementary_material}{supplementary material}} (Sec.~9).

\begin{table}[!htb]
\centering
\begin{tabular}{ | >{\centering\arraybackslash} m{10em} | >{\centering\arraybackslash} m{1cm}  | >{\centering\arraybackslash} m{1cm} | >{\centering\arraybackslash} m{1cm} | }
  \hline      
  Method                   &\small{Physical Depth} &\small{Virtual Depth}    &PCKh  \\ \hline
  Feedforward-Hourglass    &24    &-      &  77.6          \\
  Feedback-Hourglass       &4     &12     & \textbf{82.3}  \\
  \hline
\end{tabular}
\caption{\footnotesize \textbf{Evaluations on MPII Human Pose Dataset.} PCKh is the standard metric measuring body joint localization accuracy~\cite{andriluka20142d}.}
\label{table:mpii}
\end{table}

\section{Conclusion}
\vspace{-5pt}

We provided a study on feedback based learning, arguing it is a worthwhile alternative to commonly employed feedforward paradigm with several basic advantages: early prediction, taxonomy compliance, and Episodic Curriculum Learning. We also observed that the feedback based approach develops a coarse-to-fine representation that is meaningfully and considerably different from feedforward representations. 
This study suggests that it would not be far-fetched to find the useful practices of computer vision lying in a feedback based approach in the near future. \vspace{-3pt}
\\ \\
\noindent\textbf{Acknowledgement:} We gratefully acknowledge the support of ICME/NVIDIA Award (1196793-1-GWMUE), ONR (1165419-10-TDAUZ), MURI (1186514-1-TBCJE), Toyota Center (1186781-31-UDARO), and 
ONR MURI (N00014-14-1-0671).

{\small

}

\end{document}